\documentclass[runningheads,a4paper]{llncs}

\usepackage{amssymb}
\usepackage{graphicx}

\usepackage[utf8]{inputenc}
\usepackage[T1]{fontenc}

\usepackage{url}
\usepackage{hyperref}
\usepackage[margin=1.1in]{geometry}
\hypersetup{hidelinks}
\newcommand{\keywords}[1]{\par\addvspace\baselineskip
\noindent\keywordname\enspace\ignorespaces#1}

\begin{document}

\mainmatter  

\title{Opening mind by opening architecture: analysis strategies}

\titlerunning{Opening mind by opening architecture}


%
\author{Francesco Vitucci\inst{1}, Giuseppe Silvi\inst{1}, Daniele Giuseppe Annese\inst{1},\newline Francesco Scagliola\inst{1}\and Anthony Di Furia\inst{1}
}
%

\authorrunning{Francesco Vitucci et al.}



\institute{\textsuperscript{1} Conservatorio “N. Piccinni” - Bari\\
	\email{anthonydifuria.sound@gmail.com}}

%
%

\maketitle

\begin{abstract}
In numerical signal processing for electroacoustic composition, the progressive loss of specific development and research environments caused by the increasing use of digital market tools has favoured the dominance of the closed-architecture audio processor model. This model, while powerful, envisions the possibility of describing output data about its perceived characteristics, but at the cost of ignoring its internal process and interacting systems, which become complex, powerful environments but closed in an inscrutable black box, a loss we must consider. Any digital signal processing technique tells a story. Just as the words of a language incorporate social, historical and technical polysemic layers, a signal processor has its own story of implementation, a gradual technological achievement with its inevitable aesthetic consequences. Through the looking-glass of literature, one can access those environments with renewed awareness by reestablishing a scientific method and an attitude to research. In this specific case, starting from the case study of Manfred Schroeder's historical reverbs, we illustrate the process of building analytical evaluation tools, as well as practical implementation, at the basis of a conscious study path.

\end{abstract}

\section{Introduction}

The process of IT democratization started with the introduction of the Personal Computer can be observed from multiple social perspectives. Unconditional accessibility to digital signal processing tools features a wide range of signal processors (DSP) to an ever wider range audience of users. This opening has generated a new category of producer users, who, fascinated by the ease of use and the speed of obtaining results, have become accustomed to the \emph{black box} approach. This \emph{closed architecture} model represents a paradigm in which users can use a process without necessarily understanding its internal workings: they evaluate the \emph{output} based on perceived characteristics, ignoring implementation details, in one consequent
social normalization in which the user, or the music producer, «è in
realtà un \emph{music prosumer} (che ‘produce’ solo consumando servizi e dispositivi tagliati su misura)»\footnote{«is, in reality, a \emph{music prosumer} (who ‘produces’ only by consuming services and devices cut to size)», translated by authors.}. \cite{ads:circuiti} Parallel to this trend is an ever-growing lack of specific research and development environments, in an extended sense of awareness and critical consciousness. The word \emph{environment}, denotes «the conditions that you live or work in and the way that they influence how you feel or how effectively you can work»\footnote{\url{https://dictionary.cambridge.org/dictionary/english/environment}} and applies to the specific discussion, teaching or work environment, to design and implementation software, to large research centres, once the only environments where everything that is being discussed here was possible. Speaking of software, it is worth underlining that, the adoption of the model \emph{black box}, with increasingly high-level function libraries, ready for use, cannot be inspected to understand how it works, promotes a superficiality in learning and using the tools sound processing.

In this scenario, training and education play a fundamental role in promoting an \emph{open architecture} approach, encouraging students to explore and understand the theoretical and practical foundations of audio signal processing, in order to develop critical skills and creativity. This model, also called \emph{white box},  \cite{mk:wb12} predicts that the focus is on understanding the underlying mechanisms of the process, promoting critical awareness and in-depth understanding. The purpose of this discussion is therefore to show a path of understanding, acquisition and potential creative development. At the same time, we want to chart a course for future artistic and musical research works scientists who can find fertile ground in the use of the proposed paradigm to take root.

\clearpage

\section{Analysis methodology}\label{sec:anal}

The implementation path that will follow is based on a series of specific choices. First of all, we chose to observe the reverberations from the point of view of the contents because the literature in this regard is open and accessible, in line with the proposed paradigm, as well as among the richest: precisely this richness has its violent counterpart inaccessible in prosumer use. It is precisely this method of use that, for example, has led to the use of convolution reverbs, the closed application of a mathematical system that does not contain the reverberation process but only imitates one of its results. On the contrary, historical algorithmic implementations have from time to time recombined basic reverberant components to obtain different results and, precisely in this process, a possible creative outlet was identified, in continuity with the attitude of the past: these were historical moments in which implementation primarily meant dealing with the limitations imposed by the machines available, thus generating an impulse for continuous investigation.

\begin{figure}
\centering
\includegraphics[]{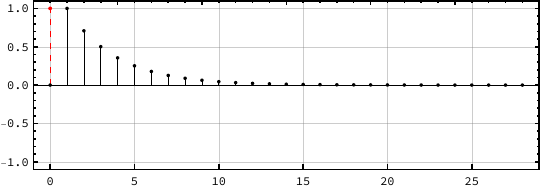}
\caption{Plot of the impulse response of the \texttt{FDL} component, obtained from within Csound with the proposed analysis routine.}
\label{fig:fdlir}
\end{figure}

The starting point of the research was the implementation of M. R. Schroeder reverberator. In 1962, he developed the first digital reverb algorithm in history \cite{schroeder1962natural}. It is made of two basic components: a \emph{delay in feedback loop}, or \emph{comb filter} and an \emph{all-pass filter}. The former is built up by inserting a simple delay line into a feedback loop; by doing so, one can produce multiple echoes, with exponential decay, as shown in Fig. \ref{fig:fdlir}. By mixing the direct sound and the delayed sound the \emph{comb filter} is converted to an \emph{all-pass filter}, a basic reverberant unit with flat frequency response.

The creation of this path was done in a textual programming environment, a choice attributable to educational needs, as the understanding of all the pieces of classical programming is more direct: declarations of variables, the definition of functions and their concatenation. In this specific case, we initially chose to use the \emph{Faust} language (\emph{Functional Audio Stream}) which is a functional programming language for sound synthesis and audio processing.\footnote{\url{https://faust.grame.fr/}} In addition to being textual, this language has a series of specific characteristics of an open architecture model: the highest level functions available are implemented using the same language, allowing the operation to be observed in every greater depth, up to the basic “primitive” components; the tools provided have analytical capabilities that allow both to visualize the sample-by-sample behaviour of a processor and to generate its block diagram. This represents the background of the “Reverberators” project\footnote{\url{https://github.com/s-e-a-m/faust-libraries/blob/master/src/seam.schroeder.lib}} \cite{rvb:faust}, a path of reconstruction and conscious implementation of reverberators from historical literature.

The idea therefore of porting the project into a Csound\footnote{\url{https://github.com/s-e-a-m/csound-libraries}} \cite{seam:csound} environment, in an attempt to preserve the chosen methodology, was immediately measured with the ability to build precise analysis tools for the impulse responses of the components under construction (Sec. \ref{sec:ir}). However, it is not simply a question of replacing a language with an alternative, but rather the integration of the various experiences, merging the strengths of the various approaches: from this perspective it is possible to observe the reconstruction of some of Faust's composition structures, such as \texttt{par} and \texttt{seq}\footnote{\url{https://faustdoc.grame.fr/manual/syntax/}}, indispensable for the implementation path (Sec. \ref{sec:parseq}). Precisely in this step, we encountered different efficiency limitations between the two languages. Therefore, understanding that these are teaching study strategies, we want to underline how the construction of compiled csound opcodes in C is the next step which, in the growth of the project, will represent a future stage of development.

\clearpage

\section{Impulse Response}\label{sec:ir}

As already explored in other research works \cite{msnsar:XXIVCIM}, once a component has been developed, it is necessary to test its impulse response (\emph{IR}). To create a precise \emph{IR} analysis routine, an appropriate orchestra of two instruments was developed, as can be seen in the code snippet below. \texttt{instr 1} is responsible for generating a Dirac pulse, testing the component (Schroeder \texttt{FDL}\footnote{The \texttt{FDL\_SCH} and \texttt{APF\_SCH} UDOs can be consulted at the following link: \newline\url{https://github.com/s-e-a-m/csound-libraries/blob/main/src/schroeder.udo}}), writing the \emph{IR} in a globally accessible table and scheduling \texttt{instr 2}. The latter plots the values in the console and a txt file; then it can be parsed to extract the desired amplitude values (for this purpose a script in Wolfram Language\footnote{\url{https://www.wolfram.com/language/}} was used, which also automates the graphic representation process (Fig. \ref{fig:fdlir}).

\bigskip

\noindent
{\it Analysis routine and plot of the impulse response}
\begin{verbatim}
<CsoundSynthesizer>
<CsOptions>
-n -d -+rtmidi=NULL -M0 -m0d
</CsOptions>
<CsInstruments>
ksmps = 32
nchnls = 2
0dbfs = 1
#include "schroeder.udo"
instr 1
  setksmps 1
  ires system_i 1, "mkdir -p PLOT"
  iT = 1;time
  iG = 1/ sqrt(2);gain
  iSR = sr;sample rate
  aDirac mpulse 1,1
  aR = 0
  iLenTable = 32
  giFDL_PLOT ftgen 2, 0, iLenTable, -2, 0
  aFDL_PLOT = FDL_SCH(aDirac, iT, iG, iSR);process to test
  aIndex phasor sr/iLenTable
  tablew aFDL_PLOT, aIndex * iLenTable, giFDL_PLOT
  schedule 2, iLenTable/sr, 1
endin
instr 2
  ftprint giFDL_PLOT
  ftsave "PLOT/1FDL_PLOT.txt", 1, giFDL_PLOT
  ires system_i 1, "wolframscript -script plot.wls"
  exitnow
endin
</CsInstruments>
<CsScore>
i1 0 100
</CsScore>
</CsoundSynthesizer>
\end{verbatim}

\section{Compositional Structures: \texttt{par} and \texttt{seq}} \label{sec:parseq}

\begin{figure}
	\centering
	\includegraphics[scale=0.65]{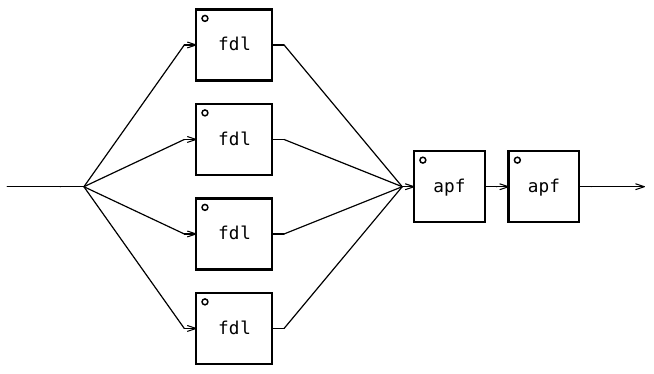}
	\caption{Faust Block diagram of Schroeder's reverb implementation: a section of four \emph{comb filters} in parallel precedes a chain of two \emph{all-pass filters} in series.}
	\label{bd:sch}
\end{figure}

Once the evaluation and analysis tools have been acquired and the actual effectiveness of the constructed basic components has been tested, the blocks can be composed. If we observe Schroeder's reverb implementation model (Fig. \ref{bd:sch}), we identify two behaviours that deserve attention: a first construct with structurally identical blocks (but with different parameters) positioned in parallel precedes a second in which the blocks are chained in series.

These two compositional structures are already supported by Faust (see Sec. \ref{sec:anal}), but are not present in canonical Csound language, so two \textit{User Defined Opcodes} were specifically written (as can be seen in the code snippet below). At the moment the proposed UDOs are not generic, as Faust \texttt{par} and \texttt{seq}; nevertheless, they are easily adaptable for any other opcode to compose in parallel and in series.

\bigskip

\noindent
{\it \texttt{par} and \texttt{seq} User Defined Opcodes}
\begin{verbatim}
;--------PAR FDL--------------------------------------------------
opcode PAR_FDL_SCH, a, ai[]i[]iii
  aPulse, iTfdl[], iGfdl[], iSR, iN, icnt  xin
	icnt  =  icnt + 1
	aFDL = FDL_SCH(aPulse, iTfdl[icnt - 1], iGfdl[icnt - 1], iSR)
	aMix init 0
	if icnt < iN then
	  aMix PAR_FDL_SCH aPulse, iTfdl, iGfdl, iSR, iN, icnt
	endif
	xout aMix + aFDL
endop
;--------SEQ APF-------------------------------------------------
opcode SEQ_APF_SCH, a, ai[]i[]iii
  aPulse, iTapf[], iGapf[], iSR, iN, icnt  xin
	icnt  =  icnt + 1
	aAPF = APF_SCH(aPulse , iTapf[icnt - 1], iGapf[icnt - 1], iSR)
	if icnt < iN then
	  aAPF SEQ_APF_SCH aAPF, iTapf, iGapf, iSR, iN, icnt
	endif
	xout aAPF
endop
\end{verbatim}

Of course, a simple chain like the one shown in Fig. \ref{bd:sch} can also be created without the illustrated UDOs. However, it becomes difficult to ignore them if the number of components in series or parallel becomes much higher, to research and study the creative expansion of historical processes, one of the topics underlying the “Reverberators” project. \cite{msnsar:XXIVCIM}

\section{Conclusions} \label{sec:conclusions}

It is appropriate to underline the strong didactic aspects of the proposed paradigm. Opening architectures means opening paths of understanding and learning; it means colliding with and circumventing external problems and limitations; it therefore means not \emph{consuming} but \emph{producing} the environment with which to interact, in a virtuous mechanism of musical, creative and professional growth.

Schroeder's work \cite{schroeder1962natural} clearly shows architectures impulse responses that led the first Faust implementation. \cite{msnsar:XXIVCIM} Its Csound porting \cite{seam:csound} produced an analysis environment, necessary for the understanding and consequent conscious writing of open architectures and further creative applications.

\end{document}